%% file: MICCAI2026-Latex-Template/MICCAI2026-main_conference_paper_template.tex
\documentclass[runningheads]{llncs}
\usepackage{hyperref}
\usepackage[T1]{fontenc}
\usepackage{amsmath,amssymb,amsfonts}
\usepackage{subcaption}
\usepackage{xspace}
\newcommand{\name}[0]{{\sc MedConcept}\xspace}
\usepackage{graphicx,verbatim}
\begin{document}
%



\title{\name: Unsupervised Concept Discovery for Interpretability in Medical VLMs}

\titlerunning{\name: Interpretability in Medical VLMs}
%
\author{Md Rakibul Haque\inst{1,2} \and
KM Arefeen Sultan \inst{1,2} \and
Tushar Kataria\inst{1,2} \and 
Shireen Elhabian \inst{1,2} 
\authorrunning{F. Author et al.}
%
\institute{Kahlert School of Computing, University of Utah\and
Scientific Computing and Imaging Institute, University of Utah
}
}

  
\maketitle              
\begin{abstract}
While medical Vision-Language models (VLMs) achieve strong performance on tasks such as tumor or organ segmentation and diagnosis prediction, their opaque latent representations limit clinical trust and the ability to explain predictions. Interpretability of these multimodal representations are therefore essential for the trustworthy clinical deployment of pretrained medical VLMs.
%
However, current interpretability methods, such as gradient- or attention-based visualizations, are often limited to specific tasks such as classification. Moreover, they do not provide concept-level explanations derived from shared pretrained representations that can be reused across downstream tasks.
%
%
We introduce \name, a framework that uncovers latent medical concepts in a fully unsupervised manner and grounds them in clinically verifiable textual semantics.
%
%
\name identifies sparse neuron-level concept activations from pretrained VLM representations and translates them into pseudo-report-style summaries, enabling physician-level inspection of internal model reasoning.
%
To address the lack of quantitative evaluation in concept-based interpretability, we introduce a quantitative semantic verification protocol that leverages an independent pretrained medical LLM 
as a frozen external evaluator to assess concept alignment with radiology reports. 
We define three concept scores, Aligned, Unaligned, and Uncertain, to quantify semantic support, contradiction, or ambiguity relative to radiology reports and use them exclusively for post hoc evaluation.
%
These scores provide a quantitative baseline for assessing interpretability in medical VLMs.
%
\href{Code}{All codes, prompt and data to be released on acceptance.}
\keywords{Unsupervised Concept Discovery  \and 3D Medical Image analysis \and Report Generation }

\end{abstract}
\input{MICCAI2026-Latex-Template/Introduction}

\input{MICCAI2026-Latex-Template/Methods}

\input{MICCAI2026-Latex-Template/Results}

\vspace{-1em}
\section{Conclusion and Future Work}
We introduce \name, a unified framework for unsupervised concept extraction and structured evaluation in medical Vision–Language Models. To the best of our knowledge, this study is among the first to systematically extract concepts from 3D volumetric foundation model representations and to formalize their quantitative verification. We apply sparse autoencoders to frozen volumetric embeddings to decompose high-dimensional features into sparse neuron-level activations, grounding them in medical terminology through a structured vocabulary within a shared vision–language embedding space. 
A central contribution is a scalable LLM-based evaluation protocol that defines Aligned, Unaligned, and Uncertain scores to quantify concept–report consistency in a reproducible and temperature-robust manner. This framework enables structured verification across cases and concept ranks, transforming concept interpretability from qualitative inspection into a measurable and comparable evaluation setting. Experiments on AbdomenAtlas 3.0 and MerlinPlus demonstrate stable scoring across decoding temperatures and consistent rank-dependent trends, highlighting robustness to LLM stochasticity.
%
Future work will focus on expanding and refining the medical vocabulary through expert curation, conducting systematic validation with clinician-annotated studies, extending evaluation to additional datasets and foundation models, and developing methods for improved concept disentanglement and hierarchical organization within volumetric representations.

%

\newpage
%
%
%
\bibliographystyle{splncs04}
\bibliography{mybibliography}
%




\end{document}

%% file: MICCAI2026-Latex-Template/Introduction.tex
\section{Introduction}
Foundation models, particularly multimodal Vision-Language Models (VLMs) trained on large-scale medical datasets, are increasingly adopted across medical image analysis tasks such as tumor segmentation, disease classification, and image-guided intervention \cite{blankemeier2024merlin,wu2025vision,alepcha2025deep}.
Despite their strong performance, these models rely on complex latent representations that remain opaque to clinicians, limiting trust and interpretability \cite{ennab2024enhancing,salahuddin2022transparency}. This opacity is especially concerning in high-stakes settings such as radiology, surgical planning, and diagnostic decision making, where erroneous or misaligned representations may lead to significant clinical consequences \cite{dhar2023challenges,d2025foundation}. 
%
%
Current interpretability techniques, including attention- and gradient-based visualizations, tend to focus on explaining individual task outputs while providing only partial visibility into the semantic organization of the underlying latent space \cite{komorowski2023towards,chung2024evaluating,fayyaz2025grad,huff2021interpretation,ennab2025advancing}. As a result, there remains a pressing need for systematic methods that can recover clinically meaningful concepts directly from pretrained representations in a task-agnostic manner.
%


Sparse autoencoders (SAEs) have emerged as a powerful tool for unsupervised concept discovery in deep neural networks, first demonstrating their impact in large language models by decomposing representations into semantically meaningful and disentangled features \cite{elhage2023monosemantic,shi-etal-2025-route,gallifant2025sparse}. Subsequent work has shown that SAEs similarly reveal sparse, interpretable structure in vision models, identifying object parts, textures, cellular structures, and clinically relevant patterns \cite{rao2024discover,bhalla2024interpreting,gong2025concepts,dasdelen2025cytosae,pach2025sparse}.
%
Building on this foundation, we introduce \name, which applies sparse decomposition to pretrained medical VLM representations to uncover clinically meaningful latent factors without manual concept annotations. \name grounds these latent representations in clinically verifiable textual medical semantics, linking neuron-level activations to semantically aligned medical concepts. The framework also produces patient-level concept summaries that clinicians can review to verify findings, refine annotations, or diagnose model errors, such as overlooked or inaccurately represented features.
The entire concept discovery pipeline operates without task-specific supervision, paired image text training, or manual concept annotations. Public medical ontologies \cite{lindberg1993unified,umls} are used to build the candidate concept vocabulary.

While concept discovery research has proposed various evaluation strategies, recent applications of SAEs in natural and medical imaging have relied primarily on manual inspection and selected visualizations rather than systematic quantitative assessment \cite{rao2024discover,bhalla2024interpreting,gong2025concepts,dasdelen2025cytosae}. Such qualitative analyses are inherently limited in scale and susceptible to selection bias, making it difficult to determine whether discovered concepts capture clinically meaningful semantics or provide a reliable basis for benchmarking models.
%
%
%
To address this gap in concept-based interpretability, we introduce a systematic LLM-based semantic entailment scoring framework for evaluating SAE-driven concept discovery in medical VLMs. Leveraging paired clinical reports and a frozen, independently pretrained medical LLM, the framework performs per-concept semantic inference to enable automated, scalable assessment of alignment between discovered concepts and textual evidence.
%
We define three complementary scores, \textit{Aligned, Unaligned}, and \textit{Uncertain}, to quantify whether predicted concepts are supported by, contradicted by, or remain ambiguous with respect to the corresponding reference radiology reports. The reports are used exclusively for post hoc evaluation, while concept discovery and extraction remain fully unsupervised. This framework enables systematic benchmarking of interpretability in current models and provides a principled foundation for evaluating future medical vision–language models.
%
%
The main contributions of this paper are as follows:

\begin{itemize}
    \item We introduce \name, an unsupervised framework for medical concept discovery from pretrained VLM representations using sparse decomposition to identify clinically meaningful neuron-level latent factors.

    \item We ground discovered latent concepts in textual medical semantics by mapping sparse activations to aligned clinical concepts and generating patient-specific concept summaries.

    \item We propose a systematic LLM-based semantic entailment protocol to perform per-concept inference on paired clinical reports, introducing three complementary metrics, \textit{Aligned, Unaligned, and Uncertain}, to quantify semantic support, contradiction, and ambiguity of discovered concepts
\end{itemize}

%% file: MICCAI2026-Latex-Template/Methods.tex
\section{\name }

\begin{figure}[!htb]
\vspace{-1.5em}
    \centering
    \includegraphics[trim={0.1cm 5.8cm 0.2cm 3.2cm}, clip=true,width=0.9\linewidth]{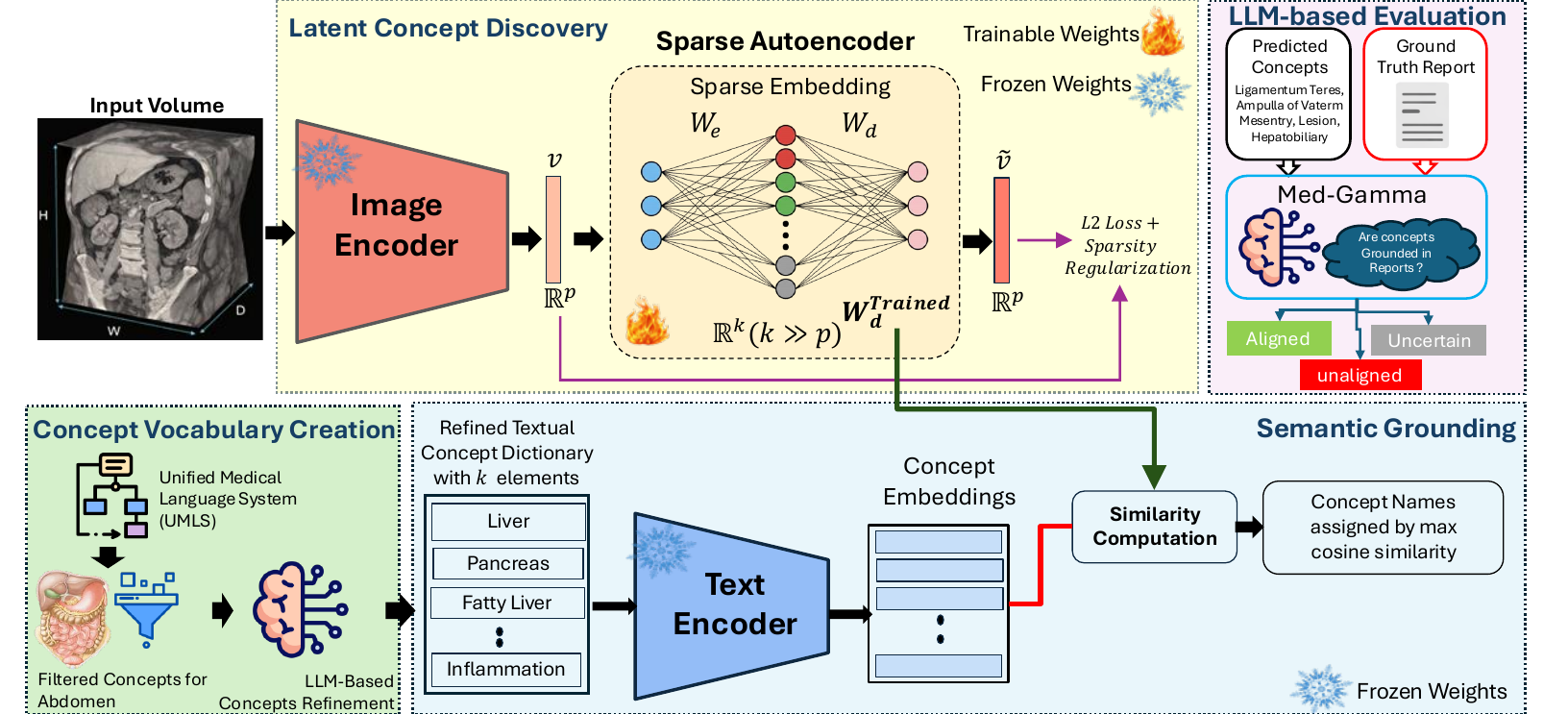}
    \caption{\textbf{\name -- and unsupervised concept discovery and semantic verification framework.} 
    A sparse autoencoder (SAE) produces sparse activations/embeddings that are matched with textual embeddings generated by a text decoder over the full concept dictionary, assigning each neuron to the concept with the highest similarity.}
    \label{fig:medsae}
    \vspace{-2em}
\end{figure}

The proposed framework comprises three sequential stages: (1) \textit{Concept discovery and extraction} via sparse autoencoder (SAE), (2) \textit{Semantic grounding} of the discovered latent concepts through alignment with textual medical concepts dictionary (shown in Figure~\ref{fig:medsae}), and (3) \textit{Quantitative evaluation} of concept–report semantic alignment.

\noindent \textbf{Latent Concept Discovery and Extraction.}
Given a volumetric image $\mathbf{x} \in \mathbb{R}^{H \times W \times D}$, where $H$, $W$, and $D$ denote the height, width, and depth (number of slices) of the volume, we extract a global feature embedding 
$\mathbf{f} = \Phi(\mathbf{x}) \in \mathbb{R}^{m}$ 
from a pretrained 3D medical vision-language model (instantiated as Merlin \cite{blankemeier2024merlin} in our experiments), where $m$ denotes the embedding dimensionality. 
Such volumetric embeddings are typically high-dimensional and poly-semantic, rendering individual dimensions neither semantically meaningful nor directly interpretable. 
To disentangle these representations into interpretable components, we train a sparse autoencoder (SAE) that maps the embedding into an over-complete sparse latent space. 
The encoder produces a sparse latent representation $\mathbf{z} \in \mathbb{R}^{k}$ and the decoder reconstructs the original embedding:
\vspace{-0.15em}
\begin{equation}
\mathbf{z} = \text{ReLU}(\mathbf{W}_{\!e} \mathbf{f} + \mathbf{b}_e), 
\qquad 
\hat{\mathbf{f}} = \mathbf{W}_{\!d} \mathbf{z} + \mathbf{b}_d.
\end{equation}
Here, $\mathbf{W}_{\!e} \in \mathbb{R}^{k \times m}$ and $\mathbf{b}_e \in \mathbb{R}^{k}$ denote the encoder weights and bias, while $\mathbf{W}_{\!d} \in \mathbb{R}^{m \times k}$ and $\mathbf{b}_d \in \mathbb{R}^{p}$ denote the decoder weights and bias. The latent dimensionality $k$ is chosen such that $k \gg m$, yielding an over-complete sparse representation. $k$ is set equal to the cardinality of the candidate concept dictionary, ensuring that each neuron in the sparse autoencoder is matched to exactly one concept.
To encourage specialized and distinct latent factors, the SAE is trained using a unified objective comprising a reconstruction term and an $\ell_1$ sparsity penalty on the activations:
\vspace{-0.15em}
\begin{equation}
\mathcal{L} = 
\underbrace{\| \mathbf{f} - \hat{\mathbf{f}} \|_2^2}_{\text{Reconstruction}}
+ 
\lambda_1 \underbrace{\| \mathbf{z} \|_1}_{\text{Sparsity}}.
\end{equation}
The sparsity term promotes activation selectivity, encouraging individual latent units to respond to specific structural patterns. 


\vspace{0.05in}
\noindent \textbf{Textual Concept Vocabulary Creation.} To create a dictionary of candidate concepts, we first extract relevant anatomical and pathological terms, along with their clinical synonyms, by querying the Unified Medical Language System (UMLS) using radiology-domain terminology. These terms are used exclusively to construct a candidate vocabulary for semantic grounding and are not used to supervise concept discovery or model training.
Because these initial UMLS queries may not capture the full lexical diversity of clinical phrasing, we subsequently employ a large language model (ChatGPT) \cite{chatgpt_2024} to systematically expand the vocabulary with additional clinically relevant synonyms, thereby curating a comprehensive corpus of abdomen-specific concepts (Figure~\ref{fig:medsae}).


\vspace{0.05in}
\noindent \textbf{Concept Naming}. An automated naming pipeline based on similarity matching is employed to align sparse autoencoder–discovered concepts with medical concepts in the candidate textual vocabulary dictionary, thereby assigning semantically interpretable labels to neuron activations.  Let $\mathcal{T} = \{t_1, t_2, \dots, t_N\}$ denote the curated vocabulary of $N$ candidate terms. Each candidate term $t_i \in \mathcal{T}$ is projected into the joint vision–language latent space using the pretrained text encoder $\psi(\cdot)$ (instantiated as Merlin in our experiments), yielding a dense embedding $\mathbf{e}_i = \psi(t_i) \in \mathbb{R}^{d}$.
To interpret the $j$-th SAE concept unit, represented by its corresponding decoder dictionary vector $\mathbf{w}_j \in \mathbb{R}^{d}$ (the $j$-th column of $\mathbf{W}_d$), we compute the cosine similarity between the latent concept representation and all candidate text embeddings:
\vspace{-0.25em}
\begin{equation}
\alpha_{ij} =
\frac{\mathbf{w}_j^\top \mathbf{e}_i}
{\|\mathbf{w}_j\|_2 \, \|\mathbf{e}_i\|_2}.
\end{equation}

The semantic label $\hat{t}_j$ assigned to the $j$-th concept unit is determined by selecting the vocabulary term that maximizes this similarity:
$\hat{t}_j = \underset{t_i \in \mathcal{T}}{\arg\max} \, \alpha_{ij}.$


\vspace{0.05in}
\noindent \textbf{Quantitative Evaluation of Concept-Report Alignment.}  The proposed framework outputs textual concepts that are highly activated for a given patient image. However, qualitative comparisons by a medical profession increases cost and is prone to observer bias and limited in scalability. To quantitatively assess semantic alignment, we leverage MedGemma \cite{medgemma} as a frozen, independent evaluator to compare the predicted concept set $\mathcal{P}$ against the paired radiology report (note that reports are used for evaluation only).

%

For a given patient image $\mathbf{x}$, we define the predicted concept set $\mathcal{P}$ as the semantic labels of all active SAE neurons:
\(
\mathcal{P} = \{\hat{t}_j \mid z_j > \tau, ~\forall j \in \{1, \cdots, k\}\},
\)
where $\tau$ denotes a predefined activation threshold. If no concept units exceed the activation threshold, the image is excluded from evaluation.
For each predicted concept $p_i \in \mathcal{P}$, MedGemma performs per-concept semantic inference conditioned on the radiology report, producing a probability distribution over three clinical states:
\vspace{-0.25em}
\begin{equation}
\mathbf{s}_i =
g_{\text{MedGemma}}(p_i, \mathcal{R})
=
\big[s_i^{c}\big]_{c \in \{\text{Aligned}, \text{Unaligned}, \text{Uncertain}\}},
\end{equation}
where $s_i^{c} \in [0,1]$ and $\sum_{c} s_i^{c} = 1$.
We obtain a discrete verdict $v_i = \arg\max_{c} s_i^{c}$ for each predicted concept and define normalized image-level alignment metrics:
\vspace{-0.25em}
\begin{equation}
\text{Score}(c)
=
\frac{1}{|\mathcal{P}|}
\sum_{i=1}^{|\mathcal{P}|}
\mathbf{1}(v_i = c),
\quad
c \in \{\text{Aligned}, \text{Unaligned}, \text{Uncertain}\}.
\end{equation}
where $\mathbf{1}(.)$ denotes the indicator function. By construction, the three scores sum to one, providing a normalized quantitative summary of semantic support, contradiction, and ambiguity relative to the radiology report.

%% file: MICCAI2026-Latex-Template/Results.tex
\section{Results}
\vspace{-0.5em}

\noindent \textbf{Datasets.}
%
We evaluate our framework on two large-scale 3D abdominal CT datasets to assess generalization across distributions and annotation types. AbdomenAtlas 3.0 \cite{abdomenatlas} includes 9,262 volumes with radiology reports and per-voxel tumor annotations from 17 public datasets. Merlin Plus contains 25,494 volumes with dense anatomical annotations generated via R-Super \cite{rsuper}.

For each dataset, we extract volumetric embeddings using the frozen Merlin encoder and train a separate sparse autoencoder (SAE) for dataset-specific concept discovery. SAE training uses only image embeddings, with radiology reports reserved for post hoc semantic evaluation.

\vspace{0.05in}
\noindent \textbf{Implementation Details.}
For each dataset, the SAE is trained on frozen feature representations extracted from the Merlin image encoder. All CT volumes are preprocessed using the same transformation pipeline as the pretrained encoder to ensure feature consistency. 
The SAE is optimized using Adam with a learning rate of $5\times10^{-5}$, an expansion factor of $2$ (i.e., latent dimensionality $k = 2p$), and an $\ell_1$ sparsity coefficient of $2\times10^{-3}$. 
Quantitative concept evaluation is performed using MedGemma-1.4B as a frozen semantic evaluator. GPT (ChatGPT v5.2) is used solely to expand and refine the UMLS-derived vocabulary during concept dictionary construction and is not involved in representation learning or evaluation.



\begin{figure}[!b]
\vspace{-1em}
    \centering
    \includegraphics[width=0.8\linewidth]{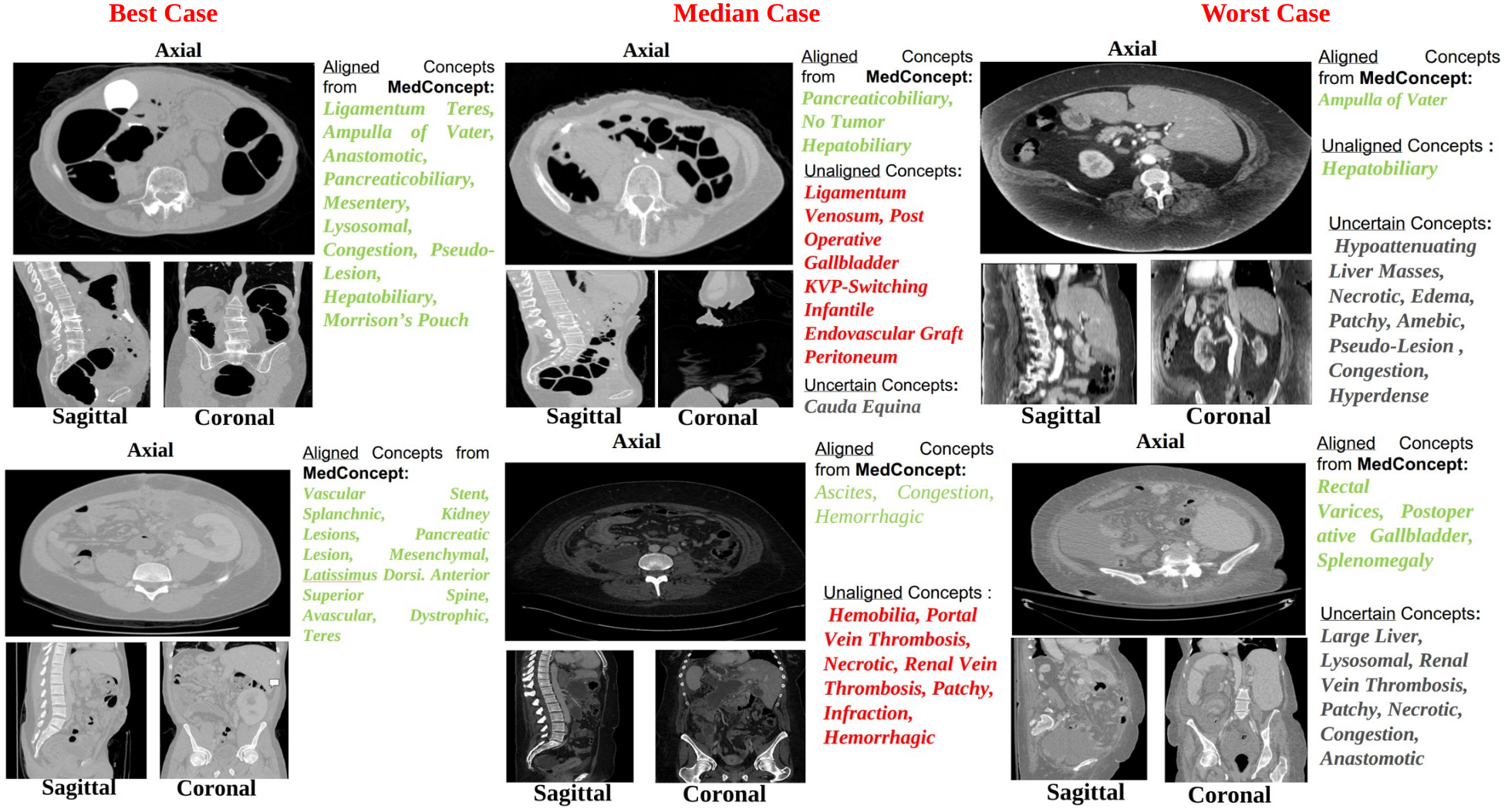}
    \vspace{-1em}
    \caption{\textbf{Qualitative evaluation of unsupervised 3D concept extraction using \name}. The top ten activated concepts per case are displayed. Concepts are evaluated post hoc for clinical grounding via \textit{MedGemma}, which classifies each concept as \textit{Present} (aligned), \textit{Absent} (unaligned or contradicted), or \textit{Uncertain}. Results indicate coherent sparse decomposition and consistent semantic grounding across datasets}
     \label{fig:3dsae_cross_dataset}
     \vspace{-1.5em}
\end{figure}

\vspace{0.05in}
\noindent \textbf{Qualitative analysis.} 
Figure~\ref{fig:3dsae_cross_dataset} presents representative qualitative examples of \name across best, median, and worst cases, ranked by the per-image Aligned score. Each case is visualized using axial, sagittal, and coronal views, alongside the top activated concepts grouped by semantic verification outcome (Aligned, Unaligned, Uncertain).
In \textit{high-scoring cases}, the majority of predicted concepts are classified as Aligned and form coherent anatomical and pathological clusters, including hepatobiliary, vascular, mesenteric, ascites, and congestion-related terms.. These thematically consistent concepts correspond to visually plausible structures in the CT volumes, consistent with structured semantic decomposition of the learned 3D representations.
In \textit{median cases}, predictions are increasingly dominated by Unaligned concepts, indicating contradictions or lack of support relative to the paired radiology report. These often involve context-dependent or overly specific descriptors such as procedural details or rare findings, suggesting semantic overprediction despite partial anatomical relevance. Importantly, because evaluation is conditioned on report evidence, Unaligned labels may reflect either model overreach or incomplete documentation within radiology reports, which frequently omit normal or secondary findings; they therefore indicate absence of textual support rather than definitive clinical incorrectness.
%
In the \textit{lowest-scoring cases}, predictions are predominantly classified as Uncertain rather than explicitly Unaligned. These concepts often involve generic imaging descriptors such as hypoattenuating, patchy, necrotic, and hyperdense, which are inherently challenging to verify when reports are incomplete or lack sufficient details. This shift toward uncertainty suggests that performance degradation is driven more by semantic ambiguity and report incompleteness than by arbitrary or anatomically implausible predictions.
%
The progression from predominantly Aligned predictions to regimes characterized by increased Unaligned (contradiction-driven) and subsequently Uncertain (ambiguity-driven) concepts mirrors the quantitative trends in verification scores. Collectively, these examples demonstrate that the framework produces structured and interpretable concepts when supported by sufficient textual evidence, while performance degradation reflects semantic overreach or report incompleteness rather than arbitrary or anatomically implausible predictions.

\begin{figure}[!b]
\vspace{-1em}
    \centering
    \includegraphics[trim={1.8cm 0.25cm 5.0cm 0.0cm}, clip=true,width=0.8\linewidth]{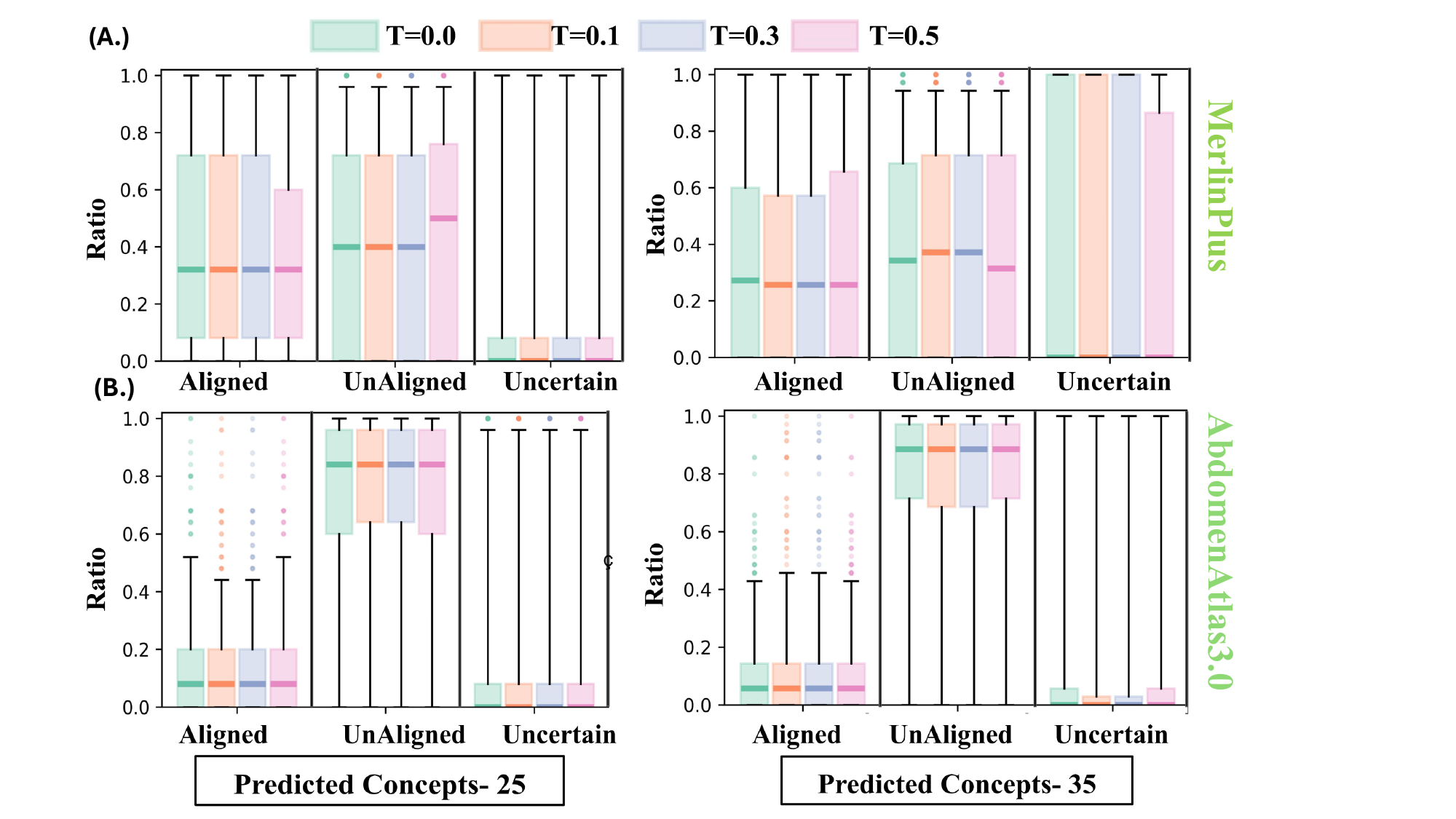}
    \caption{\textbf{Boxplots of concept verification scores for the top-$K$ predicted concepts ($K \in {25,35}$)}. Each panel shows the distribution of per-volume \textbf{Concept Alignment} (Aligned), \textbf{Concept UnAlignment} (UnAligned), and \textbf{Concept Uncertain} scores aggregated across cases. Alignment scores are consistently higher on the (A.) MerlinPlus dataset compared to (B.) Abdomen Atlas, while Abdomen Atlas exhibits comparatively lower alignment and higher unalignment proportions across temperature settings.}
     \label{fig:figure3}
\end{figure}

\vspace{0.05in}
\noindent \textbf{Quantitative Analysis:}
As shown in Fig.~\ref{fig:figure3}, concept verification scores exhibit consistent and interpretable patterns across datasets and settings. Distributions remain nearly invariant across LLM decoding temperatures from 0.0 to 0.5, indicating that the scoring procedure is stable and not driven by sampling variability under fixed prompts.
On MerlinPlus, the top 25 predicted concepts achieve a median Aligned score of approximately 0.30–0.35, while Uncertain scores typically remain below 0.1. Expanding to the top 35 concepts results in a moderate decrease in alignment and a corresponding increase in uncertainty. This rank-dependent shift is consistent with the inclusion of lower-activation units and suggests an implicit ordering within the sparse representation, where higher-activation units tend to correspond to more semantically supported concepts.
The coherent behavior across ranks indicates structured semantic organization within the learned volumetric representations and meaningful concept assignment through cosine-based retrieval over the curated medical vocabulary.
In contrast, AbdomenAtlas 3.0 exhibits substantially lower alignment, with median Aligned scores below 0.1 and Unaligned scores above 0.8 for the top 25 concepts. The top 35 configuration shows a similar distribution with only a modest increase in uncertainty. These patterns remain stable across temperature settings. The reduced alignment on this dataset is likely influenced by differences in report density and documentation style, which constrain textual evidence available for semantic verification.

\vspace{0.05in}
\noindent 
\textbf{Limitations.} While Figures~\ref{fig:3dsae_cross_dataset} and~\ref{fig:figure3} illustrate structured concept discovery without manual annotations, quantitative alignment scores are inherently bounded by the characteristics of the paired radiology reports used for evaluation. Because semantic verification is conditioned on report evidence, concepts visually present in the CT volume but omitted from documentation are penalized as Unaligned. This limitation reflects the selective and non-exhaustive nature of clinical reporting rather than definitive semantic error. 
Evaluation is further constrained by vocabulary coverage. Although curated from UMLS and refined via LLM expansion, the predefined concept set cannot exhaustively represent all abdominal pathologies, anatomical variants, or contextual findings.
Beyond textual constraints, the framework has an algorithmic limitation: semantic naming relies on cosine similarity within the joint VLM latent space and therefore inherits representational biases and potential vision–text misalignments present in the underlying pretrained model.